\documentclass[10pt,twocolumn,letterpaper]{article}

\usepackage{iccv}
\usepackage{times}
\usepackage{epsfig}
\usepackage{graphicx}
\usepackage{amsmath}
\usepackage{amssymb}

\usepackage{microtype}
\usepackage{graphicx}
\usepackage{subfig}
\usepackage{booktabs} 

\usepackage{algpseudocode}

\usepackage{amsthm}
\newtheorem{thm}{Theorem}

\newtheorem{assumption}[thm]{Assumption}
\newtheorem{remark}{Remark}

\usepackage{nicefrac}
\usepackage{comment}
\usepackage{adjustbox}
\usepackage{arydshln}
\usepackage{lastpage}
\usepackage[T1]{fontenc}

\usepackage{multirow}

\usepackage{algorithm}

\usepackage[breaklinks=true,bookmarks=false]{hyperref}

\iccvfinalcopy 

\begin{document}

\title{
Dynamic Mini-batch SGD
for Elastic Distributed Training: \\
Learning in the Limbo of Resources
}


\author{Haibin Lin, Hang Zhang, Yifei Ma, Tong He, Zhi Zhang, Sheng Zha, Mu Li \\
Amazon Web Services\\
Palo Alto, CA \\
{\tt\small \{haibilin, hzaws, yifeim, htong, zhiz, zhasheng, mli\}@amazon.com}
}

\maketitle

\begin{abstract}
With an increasing demand for training powers for deep learning algorithms and the rapid growth of computation resources in data centers, it is desirable to dynamically schedule different distributed deep learning tasks to maximize resource utilization and reduce cost.
In this process, different tasks may receive varying numbers of machines at different time, a setting we call elastic distributed training.
Despite the recent successes in large mini-batch distributed training, these methods are rarely tested in elastic distributed training environments and suffer degraded performance in our experiments, when we adjust the learning rate linearly immediately with respect to the batch size.
One difficulty we observe is that the noise in the stochastic momentum estimation is accumulated over time and will have delayed effects when the batch size changes.
We therefore propose to smoothly adjust the learning rate over time to alleviate the influence of the noisy momentum estimation.
Our experiments on image classification, object detection and semantic segmentation have demonstrated that our proposed Dynamic SGD method achieves stabilized performance when varying the number of GPUs from 8 to 128.
We also provide theoretical understanding on the optimality of linear learning rate scheduling and the effects of stochastic momentum.


\end{abstract}
\vspace{-0.8em}



\section{Introduction} 
\label{intro}

Deep learning has yet become the de facto standard algorithm in computer vision. 
Deeper and larger models keep boosting the state-of-the-art performance in image classification~\cite{he2016deep,szegedy2015going}, object detection~\cite{ren2015faster,liu2016ssd} and semantic segmentation~\cite{long2015fully,chen2018deeplab}. 
In addition, computation-intensive tasks such as video classification~\cite{wang2018non,feichtenhofer2018slowfast}, segmentation~\cite{li2018low,voigtlaender2019feelvos} and visual question answering~\cite{tapaswi2016movieqa,lei2018tvqa} bring more interests to the community as the more computation resources become available.
Despite the rapid growth of the computation powers in the data centers of research labs and leading IT companies, the demand of training deep learning models can be barely satisfied. To train a job using multiple machines, we may need to wait for a long time for the job to start due to limited resources. To finish high-priority training jobs in time, we may reserve resources in advance or stop other low-priority jobs. 

\begin{figure}
\centering
\includegraphics[width=.45\textwidth]{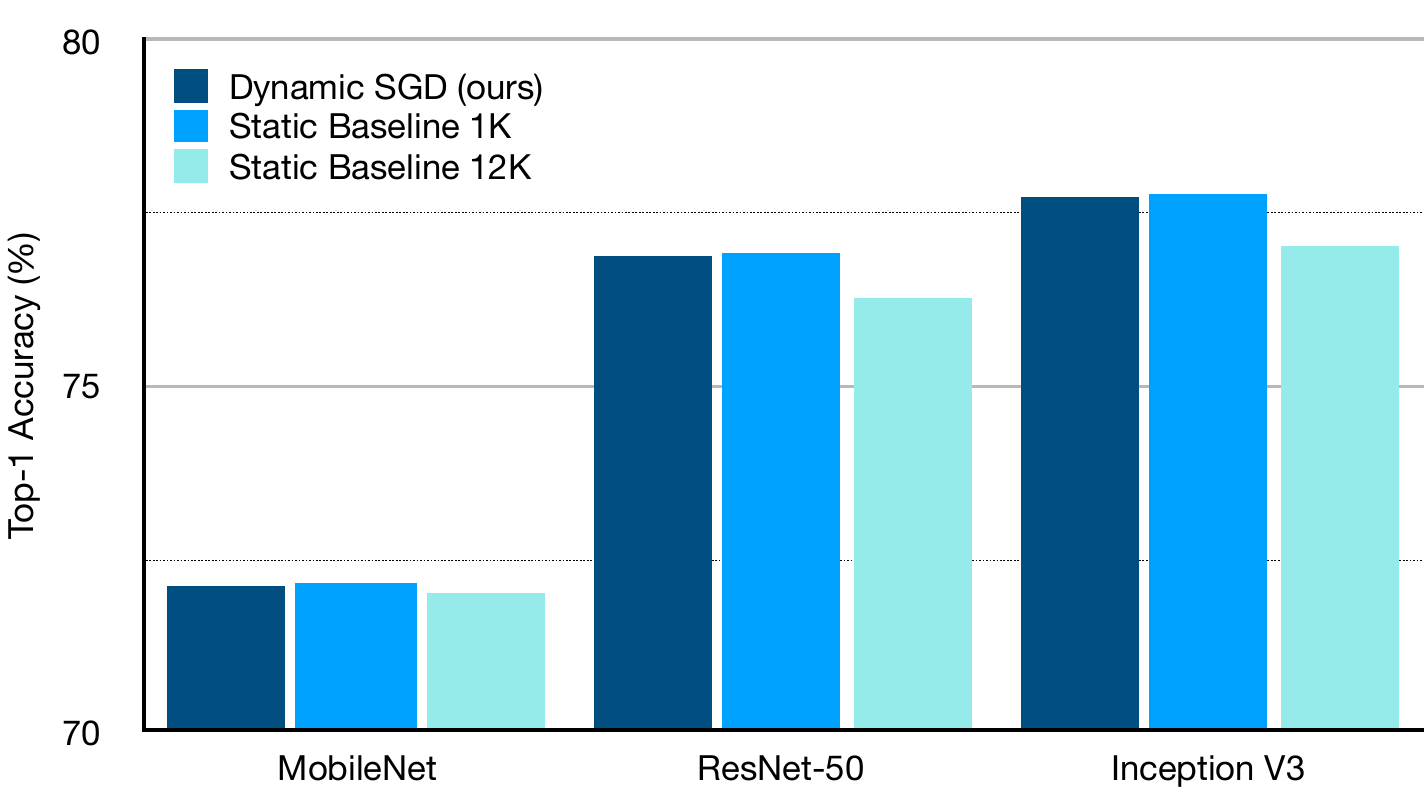}
\caption{
Top-1 accuracy on ImageNet dataset {\it vs.} different training methods. 
Static baseline refers to training with mini-batch size and number of machines fixed.
In elastic distributed training environments, the mini-batch size is dynamically updated. 
The proposed Dynamic SGD method enables stable training in such dynamic environments while benefits from the speedup with elastic resources. 
Our elastic training method achieves comparable model accuracy with small mini-batch training (1K), and always outperforms the large mini-batch one (12K) under the same setting. 
}
\label{fig:result}
\end{figure}

Allowing resources for a training job to change dynamically can greatly reduce the waiting time and improve resource utilization. We could start a job earlier when a portion of the requested resources is ready. We could also preempt resources from running low-priority training jobs without stopping them for high-priority tasks. In addition, cloud providers encourage users to use preemptible resources with significant lower prices, such as spot instance in AWS~\cite{ec2spot}, low-priority virtual machines in Azure~\cite{azure} and preemptible virtual machines on Google Cloud~\cite{google}. We refer to the above dynamic machine environment where the machines could be added or removed from a task 
as an elastic distributed training environment.

Mini-batch stochastic gradient descent (SGD) with momentum (e.g. heavy-ball momentum, Nesterov momentum, Adam~\cite{nesterov1983method, kingma2014adam}) is a widely used optimization method to train deep learning models in computer vision. Previous work focuses on asynchronous SGD (asyncSGD) for environments where machines are loosely coupled~\cite{NIPS2012_4687}. AsyncSGD, however, converges radically differently from synchronous mini-batch SGD~\cite{zhang2015staleness, reddi2015variance, zheng2017asynchronous}. It is also difficult for asyncSGD to match the model accuracy trained with synchronized mini-batch SGD with similar computation budget~\cite{NIPS2012_4687, chen2016revisiting}. 

In this work, we focus on extending synchronized mini-batch SGD with momentum into the dynamic training environment. We aim to minimize the convergence difference when the training environment is constantly changing.
There are two straightforward approaches to extend synchronized SGD with momentum. One approach is to fix the mini-batch size but reassign mini-batches across machines when the number of machines changes. With the mini-batch size fixed, the number of machines has little impact on the optimization method, but reduces the computational efficiency when adding many machines, since the assigned per machine mini-batch size decreases linearly. The other approach fixes the mini-batch size per machine, and updates the total mini-batch size linearly with the number of machines, while linearly changing the learning rate at the same time~\cite{goyal2017accurate, devarakonda2017adabatch, smith2017don}.
Empirically, we find this approach often leads to degraded model accuracy when the number of machines changes dramatically. 
Based on our theoretical analysis, the noise in the momentum state scales inversely with the mini-batch size. Linearly scaling the learning rate when increasing mini-batch size fails to consider 
the difference of variance noise scale, 
which leads to the optimization difficulty.

In this paper, we propose a new optimization strategy called {\it Dynamic Mini-batch Stochastic Gradient Descent (Dynamic SGD)} that smoothly adjusts the learning rate to stabilize the variance changes. We design multiple dynamic training environment settings by varying the number of GPUs between 8 and 128. We evaluate our proposed method on image classification (ResNet-50~\cite{he2016deep}, MobileNet ~\cite{howard2017mobilenets} and InceptionV3~\cite{DBLP:journals/corr/SzegedyVISW15} on ImageNet~\cite{deng2009imagenet}), object detection (SSD~\cite{liu2016ssd} on MS-COCO~\cite{lin2014microsoft}), and semantic segmentation (FCN~\cite{long2015fully} on ADE20K~\cite{zhou2017scene}). The experiment results demonstrate that the convergence of Dynamic SGD in dynamic environments consistently matches the convergence with training in static environments.

The contributions of this paper include:
\begin{enumerate}
\item Identifying the key challenge to extend synchronized SGD with momentum into dynamic training environments.
\item Proposing a new method which we call Dynamic SGD to smoothly adjust the learning rate when the number of machines changes to stabilize the training. 
\item Extensive empirical studies to benchmark the proposed Dynamic SGD method with straightforward approaches on large-scale image classification, object detection and semantic segmentation tasks. 
\end{enumerate}


\section{Background}


\subsection{Mini-batch SGD with Momentum}

We first review mini-batch stochastic gradient descent (SGD), and SGD with momentum update~\cite{robbins1951stochastic}. 
Given a network parameterized by vector $w$ and a labeled dataset $X$, the loss to minimize can be written in the following form~\cite{goyal2017accurate}:
\begin{equation}
L(w) = \frac{1}{|X|} \sum_{i=1}^{|X|} l(w, x_i) ,
\end{equation}

where $l(w, x_i)$ is the loss for the sample $x_i$ with parameters $w$.
SGD iteratively updates parameters $w$ with its gradient estimated within each mini-batch to optimize the loss: 
\begin{equation}
\label{eq:sgd}
w_{t+1} = w_t - \eta \frac{1}{B} \sum_{i=1}^{B} \nabla l(w_{t}, x_i) ,
\end{equation}

where $w_t$ is the network parameter at iteration $t$, $\eta$ is the learning rate, and $B$ is the number of samples  randomly drawn from dataset $X$ in a mini-batch.

In practice, SGD with momentum helps accelerate the optimization. The momentum state keeps the exponentially weighted past gradient estimates and updates the loss with the following rule:
\begin{equation}
\label{eq:momSGD}
\begin{split}
&u_{t+1} = \mu u_t + \frac{1}{B} \sum_{i=1}^{B} \nabla l(w_{t}, x_i)  \\
&w_{t+1} = w_t - \eta u_{t+1} ,
\end{split}
\end{equation}

where $u_t$ is the momentum state of historical gradients at iteration $t$, and $\mu$ is the decay ratio for the momentum state. 
The parameter update depends on both gradient estimated within current mini-batch and exponentially weighted gradients from historical mini-batches.


\subsection{Data Parallel Distributed Training}

\begin{algorithm}[t!]
\caption{Data parallel distributed training for a single mini-batch.}
\label{algo:data_para}
    \begin{algorithmic}
    \Require Mini-batch size $B$, number of workers $N$, and learning rate $\eta$
    \State Randomly sample $B$ inputs $x_1, \ldots, x_B$
    \State Partition inputs into $N$ parts $X_1, \ldots, X_N$
    \For{worker $i = 1, \ldots, N$}\Comment{Run in parallel}
       \State Compute gradient $\nabla l_i$ based on data partition $X_i$
       \State Allreduce gradient by $\nabla l \gets \sum_{i=1}^N \nabla l_i$
       \State Update parameters by $w \gets w - \frac\eta B \nabla l$
    \EndFor
    \end{algorithmic}
\end{algorithm}

In distributed training, multiple workers work together to finish a training job. A worker is a computational unit, such as a CPU or a GPU. Data parallelism defines how training workloads are partitioned into each worker.

Assume we are using mini-batch SGD with a mini-batch size $B$ on $N$ workers, and each worker has a copy of model parameters. In data parallelism, we partition a mini-batch into $N$ parts, so 
each worker will get 
$B/N$ examples, and then compute its local gradients. 
All local gradients are averaged over workers through synchronous communication to obtain the global gradient for this mini-batch. Then this global gradient is used to update model parameters by the SGD update rule. Algorithm~\ref{algo:data_para} illustrates how a single batch is computed.

\section{Methods}
\label{methods}


In this section, we first describe the setup of elastic distributed training environment, and then study the optimization instability of momentum SGD in such environment and introduce Dynamic Mini-batch SGD to stabilize the training.

\subsection{Elastic Distributed Training System}

In a dynamic scheduling deep learning system, the computation resources are managed, planned and distributed dynamically 
for different training tasks based on their priorities and system availability. 
We refer to the distributed training environment with dynamical computation resources as {\it Elatsic Distributed Training}. 

\begin{figure}
\centering
\includegraphics[height=3.4in, width=3in]{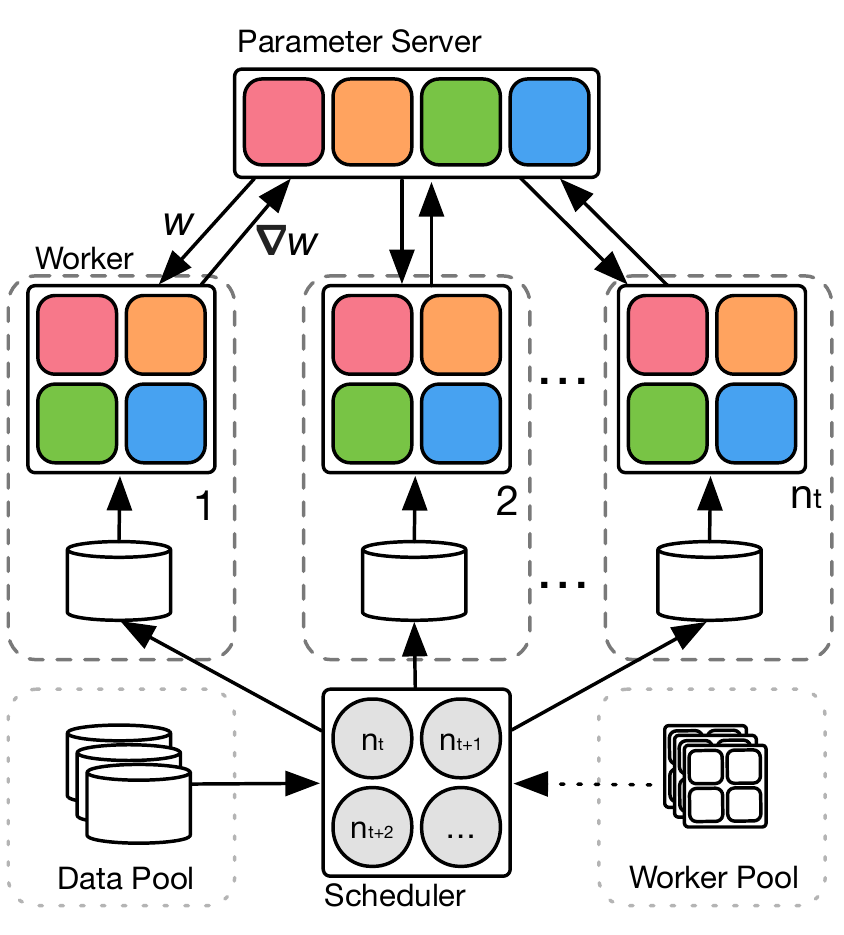}
\caption{Overview of the elastic distributed training system with an example task.
The scheduler keeps track of the training progress, monitors the worker pool, and dynamically assigns the data and workers to each task based on their priority and system availability. 
The task parameter server aggregates the parameter gradient $\nabla w$  from each worker and send back the updated parameter weights $w$. 
($n_t$, $n_{t+1}$, and $n_{t+2}$ represent the number of workers assigned to the current task at time step $t$, $t+1$ and $t+2$.)}
\label{fig:system}
\end{figure}

An overview of elastic distributed training diagram for an example task is shown in Figure~\ref{fig:system} (note that the system typically has more than one training tasks). 
The scheduler maintains the training state and coordinates the resources. It tracks the current and future number of workers by monitoring heartbeats from existing workers and availability notices.
Each worker computes the gradients for the assigned data. 
The parameter server for the training task maintains the primary copy of model parameters,  updates the parameters based on the aggregated 
gradients from workers, and send the updated parameters back to workers. 
In practice, the parameter server can 
partition the parameters among multiple machines to increase throughput.

\subsection{Dynamic Mini-batch SGD} 
\label{sec:mini-batch-sgd}

\paragraph{Learning Rate Scaling. }

Prior work linearly scales the learning rate according to the mini-batch size, which achieves success in large mini-batch training~\cite{goyal2017accurate}.
Considering the SGD without momentum update in Equation~\ref{eq:sgd},
the parameter update for $k$ iterations using mini-batch size of $B$ can be approximated by updating once with linearly scaled learning rate $k\eta$ on the combined $k$ mini-batches with the size of $kB$, if we could assume $l(w_t, x) \approx l(w_{t+j}, x) \forall j<k$. We discuss in further detail in the Appendix~\ref{app:1}. We refer to this strategy of linear scaling up the learning rate as {\it linear scaling}. 
Despite its success in large mini-batch training, linear scaling fails to apply to an elastic training environment, because the effect of momentum is not compensated. To our knowledge, the effect of momentum for changing mini-batch size has not been studied in the previous work.

\paragraph{SGD Momentum with Changing Mini-batch Size. }

To clearly see the weight update in momentum SGD as in Equation~\ref{eq:momSGD}, a common way to rewrite the equation is substituting $\eta u_t$ with $v_t$ to absorb the learning rate $\eta$~\cite{goyal2017accurate}: 
\begin{equation}
\label{eq:mom1}
\begin{split}
    &v_{t+1} = \mu v_t + \eta \frac{1}{B}\sum_{i=1}^B \nabla l(w_{t}, x_i)  \\
    &w_{t+1} = w_t - v_{t+1} ,
\end{split}
\end{equation}

which is identical to Equation~\ref{eq:momSGD} for a static learning rate $\eta$ and mini-batch size $B$. 
For elastic training at iteration $t+1$, the mini-batch size is changed from $B$ to $kB$ ($k$ is the changing ratio, which  can be decimal). Applying the linear scaling directly for learning rate to the momentum SGD, we can get the paramter update as:
\begin{equation}
\label{eq:mom2}
\begin{split}
    &v_{t+1} = \mu k v_t^B + k\eta \frac{1}{kB}\sum_{i=1}^{kB} \nabla l(w_{t}, x_i)  \\
    &w_{t+1} = w_t - v_{t+1} ,
\end{split}
\end{equation}

where $v_t^B$ is the momentum state estimated on previous mini-batch size $B$, which is scaled to $k$ times for momentum correction to maintain the equivalence with Equation~\ref{eq:momSGD}~\cite{goyal2017accurate}. 
However, this compensation only considers the gradient scale\footnote{The scale of the momentum state is $\frac{1}{1-\mu}$ times of the gradient scale.} instead of the noise scale in the momentum state.

\begin{figure}
\centering
\includegraphics[width=0.8\linewidth]{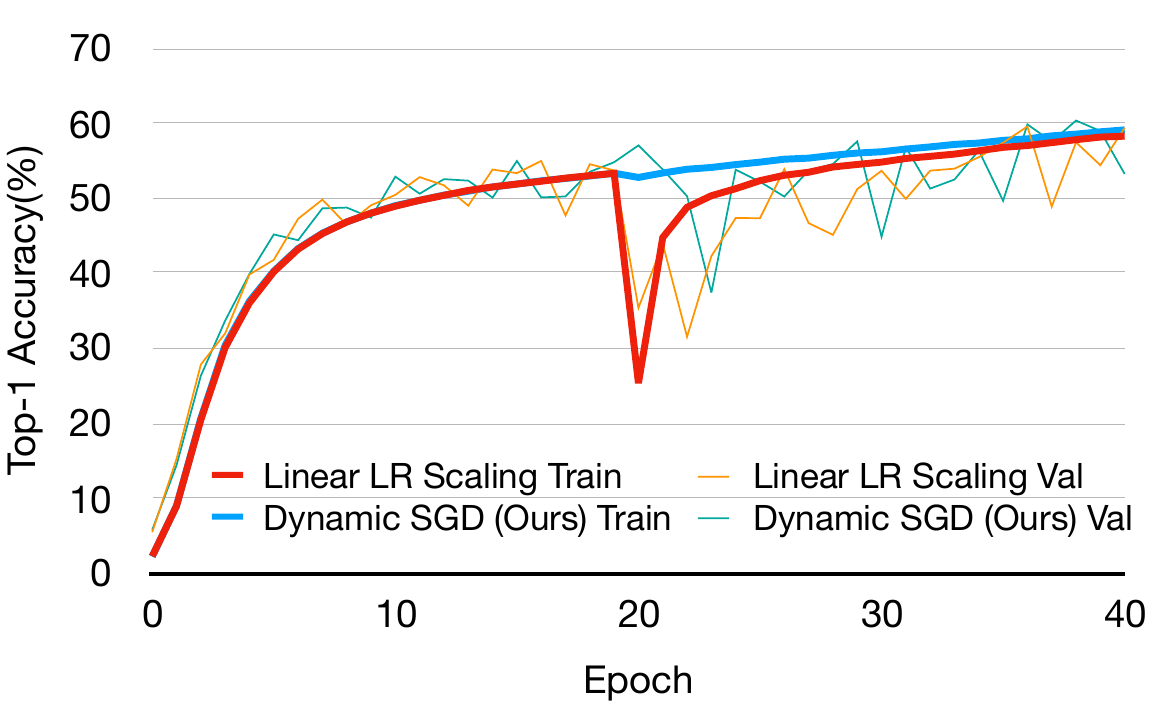}
\caption{Training and validation
accuracy on ImageNet. Increasing the mini-batch size from 1K to 12K at epoch 20. The proposed Dynamic SGD is robust to the mini-batch change, while the training with linear scaling get degraded.}
\label{fig:spike_compare}
\end{figure}

\paragraph{Noise Scale in the Gradient and Momentum State. }

SGD obtains an estimated gradient within a mini-batch of size $B$ with $\nabla l^B(w)=\frac{1}{B}\sum_{i=1}^B \nabla l(w, x_i)$ to approximate the full gradient on the entire dataset $\nabla L(w)$. 
The gradient of each sample $\nabla l(w, x_i)$ is a random variable whose expected value is $\nabla L(w)$. The variance of estimated gradient within a mini-batch scales inversely to the mini-batch size $B$~\cite{DBLP:journals/corr/abs-1812-06162}\footnote{Assuming $x_i$ is sampled independently from the dataset and $B\ll |X|$.}. 
The gradient in a mini-batch gives a noisy estimate of the full gradient, and larger mini-batch size provides less noisy estimate. 

The variance of the momentum state is proportional to the variance of the mini-batch gradient, {\it i.e.} $var(u_t)=\frac{1}{1-\mu^2} var(\nabla l^B(w))$. 
Therefore, directly scaling up the momentum state on the mini-batch size $B$ for ``momentum correction'' as in Equation~\ref{eq:mom2} increases the noise scale quadratically to $k^2$ times of the expected momentum state on mini-batch size $kB$. 
We also observe unstable training under such setting in the experiment, and the training curves are shown in Figure~\ref{fig:spike_compare}.

\paragraph{Momentum Compensation. }

To address the difficulty of adapting previous momentum state, we introduce momentum compensation factor $\gamma_t$, which gradually changes over iterations to allow smooth adaptation of the momentum state when increasing mini-batch size to $k$ times. The weight update is given by $w_{t+1} = w_t - \gamma_{t+1} \eta u_{t+1}$, where $\gamma_{t+1}$ is:
\begin{equation}
    \gamma_{t} = \left\{ 
    \begin{aligned}
        & 1 + \frac{t-t_0}{T}(k-1) &\text{ if } (t-t_0)<T \\
        & k &\text{ otherwise}
    \end{aligned}
    \right.
\end{equation}

$t_0$ is the iteration index when the mini-batch size is changed, and $T$ is the total compensation iteration number, which we find $T=8k$ works well\footnote{We only compensate for momentum adaption when increasing mini-batch size, because reducing noise scale will not influence the training (as shown in Figure~\ref{fig:spike_dump}). }.  
We refer to the mini-batch SGD adopting momentum compensation strategy for elastic distributed  training as {\it Dynamic Mini-batch Stochastic Gradient Descent (Dynamic SGD)}\footnote{We discuss other potential momentum compensation methods in the Future Work Section~\ref{sec:conclusion}.}. 
The proposed Dynamic SGD stabilizes training as shown in Figure~\ref{fig:spike_compare}.

\section{Related Work} \label{related}
\label{sec:related}

\subsection{Large Mini-batch Distributed Training}

Prior work achieves great success in large mini-batch data parallel training for deep convolutional neural networks. 
Li~\cite{li2017scaling} shows distributed training with up to 5K mini-batch size without a loss in accuracy on ImageNet. 
Goyal {\it et al.}~\cite{DBLP:journals/corr/GoyalDGNWKTJH17} employs linear learning rate scaling rule with warm-up scheme to train ResNet-50 for image classification with large batch size up to 8K without reducing accuracy. 
Layer-wise Adaptive Rate Scaling (LARS) optimization algorithm  overcomes the optimization difficulties of larger batch training beyond 8K batch size, and scales ResNet-50 training up to 32K mini-batch. 
The mini-batch size is further increased to 64K using mixed precision training~\cite{you2017large, jia2018highly}. 
Square root learning rate scaling and longer training time for 
is also proposed for training a large mini-batch size~\cite{hoffer2017train}.
Despite their success in large mini-batch training, network training using dynamic mini-batch size is rarely studied.

\subsection{Stochastic Gradient Descent}

Asynchronous stochastic gradient descent (asyncSGD)~\cite{NIPS2012_4687} assumes machines are loosely coupled and thus suits this dynamic machine environment.
Previous study demonstrates that it is difficult for asyncSGD to match the model accuracy using synchronized SGD with similar computation cost~\cite{NIPS2012_4687, chen2016revisiting}. 
Therefore we use synchronized SGD in this work instead of async SGD to achieve better model accuracy.


Our work also benefits from pioneering studies on learning rate and mini-batch size. 
McCandlish {\it et al.}~\cite{DBLP:journals/corr/abs-1812-06162} analyzes largest useful mini-batch size based on gradient noise scale. 
Prior work proposes to increase the mini-batch size instead of decaying the learning rate during the training results in less than 1\% loss in accuracy on ImageNet when scaling learning rate up to 3 times~\cite{smith2017don,devarakonda2017adabatch}. 
Jastrz{\k{e}}bski {\it et al.}~\cite{jastrzkebski2017three} shows learning rate schedules can be replaced with batch size schedules from theoretical analysis.  
Despite changing mini-batch sizes are used in these work, reserved computation resources are required due to fixed resource schedule instead of dynamically planned. 


\section{Experimental Results}
\label{experiment}

In this section, we conduct a comprehensive benchmark of proposed Dynamic SGD and baseline approaches on image classification, object detection and semantic segmentation. 
For image classification, we compare Dynamic SGD with static baseline and linear scaling for state-of-the-art network architectures ResNet-50~\cite{he2016deep}, MobileNet~\cite{howard2017mobilenets} and the InceptionV3~\cite{DBLP:journals/corr/SzegedyVISW15} on ImageNet~\cite{deng2009imagenet}.
Then we go beyond image classification task and evaluate the proposed method for 
object detection using Single Shot multi-box Object Detector (SSD)~\cite{liu2016ssd} on MS-COCO dataset~\cite{lin2014microsoft}, and semantic segmentation using Fully Convolutional Networks (FCN)~\cite{long2015fully} on ADE20K~\cite{zhou2017scene}.

\paragraph{Baseline Approaches. }
In this experiment, we mainly compare the proposed Dynamic SGD with the following baselines:

\begin{itemize}
    \item {\it Static Baseline}: no elasticity. The number of workers is fixed during the training.
    \item {\it Fixed Mini-batch Size}: fix the mini-batch size and (re)distribute the workload evenly to available workers. 
    \item {\it Linear Scaling}: linearly scale up the mini-batch size and learning rate based on number of available workers. 
\end{itemize}

\subsection{Image Classification}
\label{sec:exp_img}

We first briefly describe the implementation details of the baseline network and the elastic training simulation. 
Then we compare the Dynamic SGD method with baseline approaches with suddenly increased number of workers using ResNet-50. 
Finally, we conduct a comprehensive study on state-of-the-art image classification models 
using randomly changing number of GPUs. 

\paragraph{Implementation Detail. }

We adopt ResNet-50~\cite{he2016deep}, MobileNet 1.0~\cite{howard2017mobilenets} and Inception V3~\cite{DBLP:journals/corr/SzegedyVISW15} as the baseline models and evaluate the performance on ImageNet-2012~\cite{deng2009imagenet} dataset. The model is implemented in GluonCV\footnote{https://github.com/dmlc/gluon-cv} with MXNet~\cite{chen2015mxnet}. Each network is trained for 90 epochs using cosine learning rate decay~\cite{DBLP:journals/corr/LoshchilovH16a}.
The learning rate is warmed-up for 5 epochs~\cite{goyal2017accurate}. We use stochastic gradient descent (SGD) optimizer and set the momentum as 0.9 and weight decay as 0.0001. We use 8 GPUs with 128 per-GPU mini-batch size as the baseline, and set the base learning rate as 0.4. We linearly scale up the learning rate when increasing the mini-batch size. The input size is 224 by 224 for ResNet-50 and MobileNet 1.0, and 299 by 299 for Inception V3. We do not apply weight decay to biases as well as $\gamma$ and $\beta$ in batch normalization layers ~\cite{jia2018highly, xie2018bag}, and will discuss its effect in the appendix.
To study convergence when the mini-batch size changes, we simulate the gradient of a large mini-batch by accumulating gradients from multiple small mini-batches before applying the parameter update. For throughput analysis, we run the training job with multiple physical machines.

\paragraph{Increasing {\it\textbf{vs}.}~Decreasing Mini-batch Size Influence. }

\begin{figure}
\centering
\subfloat[Increase the mini-batch size to 12 times at epoch 20.]{
    \includegraphics[width=0.9\linewidth]{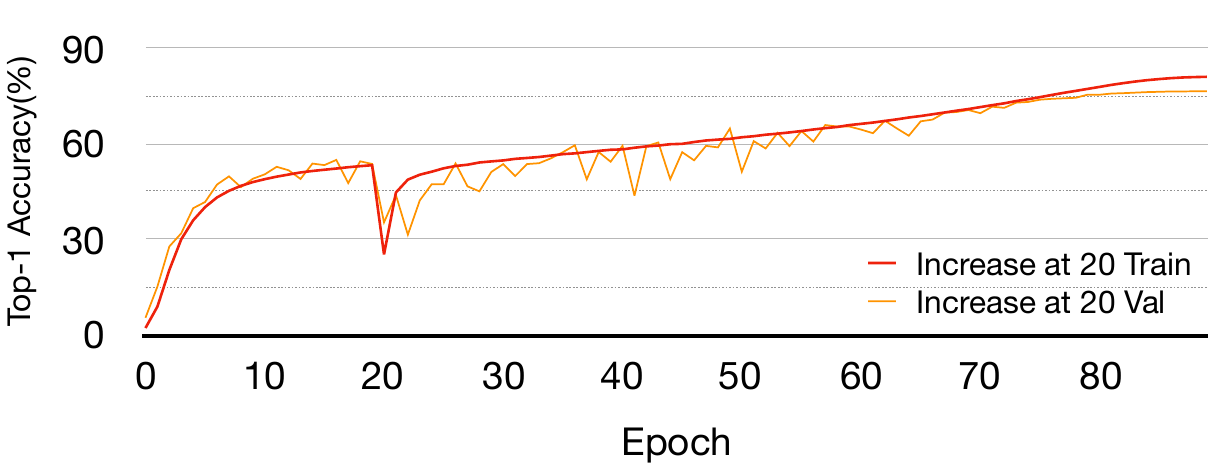}
}\\
\subfloat[Decrease the mini-batch size to $\frac{1}{12}$ at epoch 20.]{
    \includegraphics[width=0.9\linewidth]{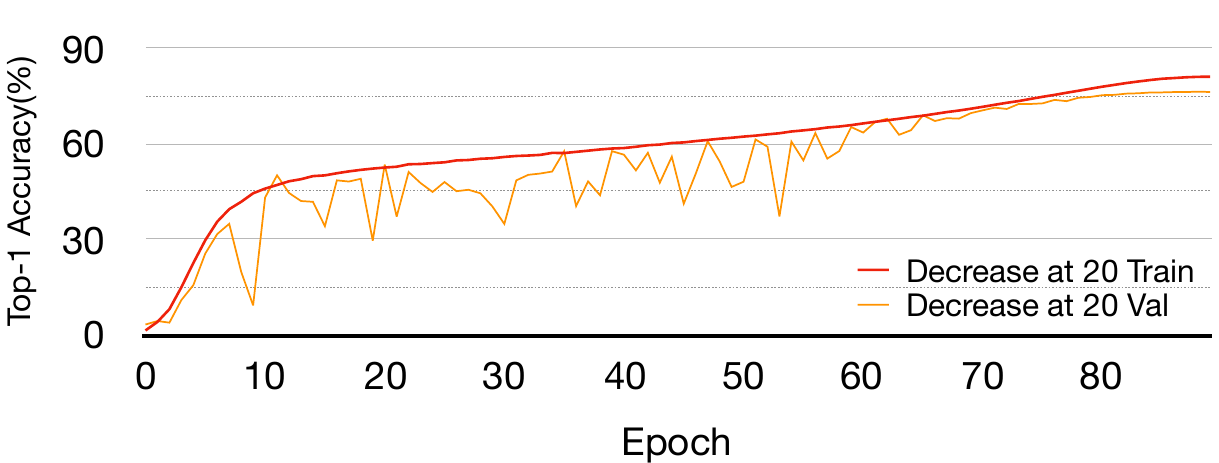}
}
\caption{Top-1 accuracy on ImageNet validation set using ResNet-50. 
Influence of sudden change in mini-batch size using linear scaling approach. 
We use the 1K as the base mini-batch size. We find that the network training is relatively sensitive to increasing mini-batch size, but robust to decreasing.}
\label{fig:spike_dump}
\end{figure}

First we study the influence of a sudden change of number of GPUs at epoch 20. We train the network with two configurations: 1) training starts from 8 GPUs and then increase the number of GPUs to 96 at epoch 20, and 2) training starts from 96 GPUs and then decrease the number of GPUs to 8 at epoch 20. In both configurations we scale up or down the learning rate linearly with the number of GPUs. The results are included in Figure~\ref{fig:spike_dump}. 
Figure~\ref{fig:spike_dump} shows that by increasing the number of GPUs to 96, the training curve has a sharp drop, indicating
the training process is drastically disrupted at epoch 20. On the other hand, decreasing the number of GPUs has no visible effect. 




\paragraph{Baseline Comparisons. }
We then focus on the scenario with increasing number of GPUs. The number of GPUs is increased from 8 to 96 at epoch 20 and epoch 70. 
Besides linear scaling, we test 1) fixing the mini-batch size when increasing GPUs, i.e. reduce the per-GPU batch size, 
and 2) incorporating the learning rate warm up after the change. The results are in Table~\ref{tab:spike_12}.

\begin{table}
\begin{center}
\begin{tabular}{l|l|l|l}
    \hline
     Method   & Scale & Epoch 20 & Epoch 70  \\ \hline\hline
     Static Baseline & 1$\times$  & \multicolumn{2}{c}{76.91} \\ 
     Static Baseline & 12$\times$ & \multicolumn{2}{c}{76.34} \\ \hline
     Fixed Mini-batch Size & 12$\times$ & 76.28 & 75.87 \\ 
     Linear Scaling   & 12$\times$ & 76.50 & 76.54 \\ 
     Dynamic SGD (ours)   & 12$\times$ & 76.81 & 76.81 \\ \hline

\end{tabular}
\end{center}
\caption{Performance from different methods when increasing the number of GPUs at early or late stage of the training process.
         Our method is usually not affected by the sudden increment and outperforms the other approaches. 
         The fixed mini-batch size method has worse performance with a late stage change.
         }
\label{tab:spike_12}
\end{table}

Table~\ref{tab:spike_12} shows that our Dynamic SGD method has a consistent performance 
against the sudden change at both early and late stage. 
Fixing the mini-batch size gives interesting results: it has a reasonable accuracy if the sudden 
increase happens at epoch 20, and gets worse if happens at epoch 70.
When keeping the mini-batch size and increasing the number of GPUs, the per-GPU batch size is going to be a smaller number,
which may affect the behavior of batch normalization layers \footnote{We discuss batch normalization in the appendix.}. Increasing at epoch 20 offers the network a longer time to
train the batch norm layers thus results in a better accuracy.

\begin{figure}
\centering
\includegraphics[width=.5\textwidth]{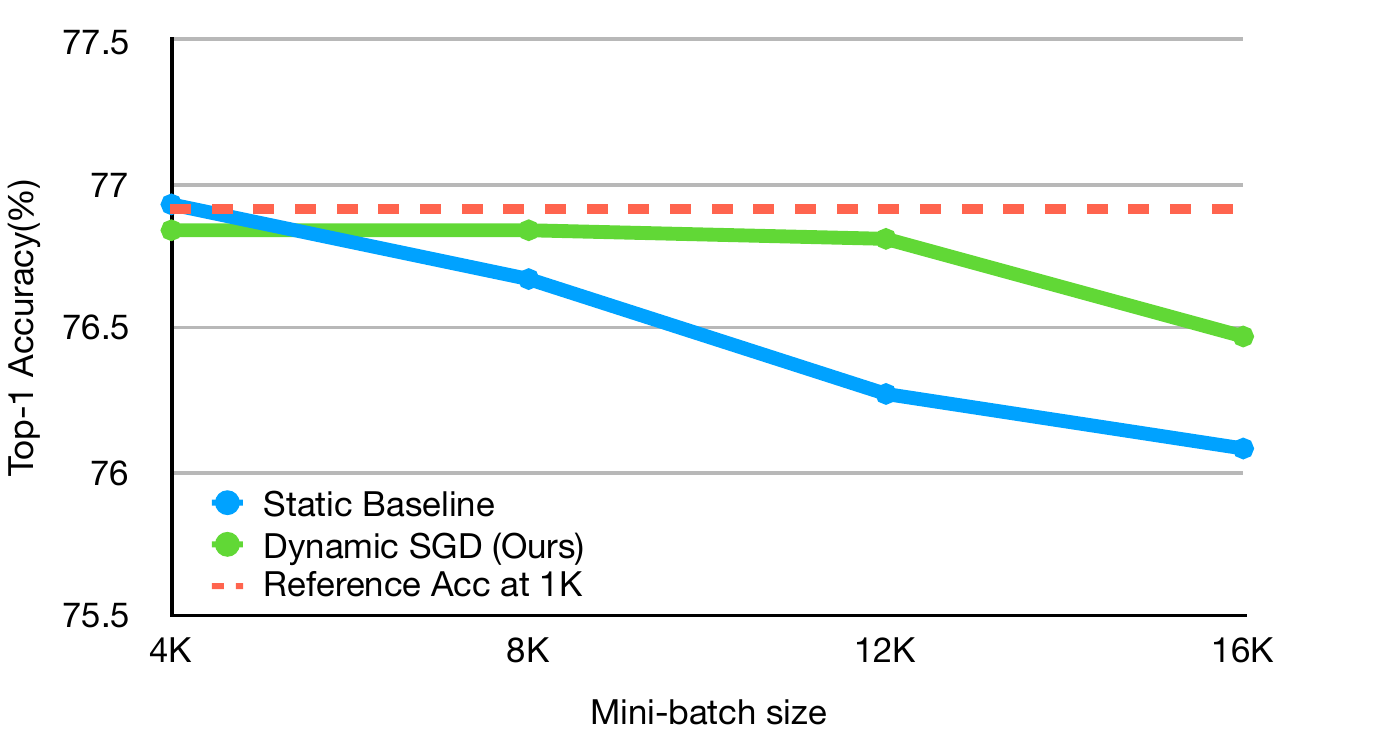}
\caption{Top-1 validation accuracy {\it vs.} mini-batch size for Dynamic SGD
         and static baseline. 
         The number of GPUs is increased at epoch 20. 
         Dotted line is the performance of static baseline at 1K. Our method
         is close to the 1K reference and is often better than the static baseline 
         at large mini-batch size.}
\label{fig:batchsize}
\end{figure}

\paragraph{Mini-batch Size Range. }
We compare our method with static baselines. By setting mini-batch size at 1K as the reference,
Figure~\ref{fig:batchsize} includes the static baselines with larger mini-batch sizes, and our method with
the mini-batch size increased at epoch 20.
The static baseline with mini-batch size at 4K is slightly better than the baseline at 1K,
which is likely to be caused by random variation. Starting from
8K the baseline keeps getting worse, while the Dynamic SGD method keeps up a better
accuracy. It means that our method preserves the better initial status 
when training with a smaller mini-batch size, and can successfully transfer it to the training
with larger mini-batch size. Our method has similar performance from mini-batch size at 4K to 12K,
and it gets worse at 16K. This may be that the assumption $l(w_t, x) \approx l(w_{t+j}, x)$ 
in Section~\ref{sec:mini-batch-sgd} does not hold anymore.

\paragraph{Throughput Analysis.}

\begin{table}
\begin{center}
\begin{tabular}{l|l|l}
    \hline
     Method   & Scale & Throughput  \\ \hline\hline
     Static Baseline & 1$\times$  & 4944 \\ \hline
     Fixed Mini-batch Size & 12$\times$ & 335 \\
     Dynamic SGD (ours) & 12$\times$ & 40095 \\ \hline
\end{tabular}
\end{center}
\caption{The training throughput (samples/sec) when increasing the number of GPUs from 8 to 96 at epoch 20. The static baseline uses a fixed number of GPUs of 8. The training throughput increases to 8.11 times when the number of GPUs increases by 12 times with Dynamic SGD method. For the fixed mini-batch size method, we observe significant slowdown when the number of GPUs increases.}
\label{tab:throughput}
\end{table}

We benchmark the training throughput of both fixed mini-batch method and the Dynamic SGD method. We use 12 machines, each equipped with 8 V100 GPUs and 2 Intel Xeon E5-2686 CPUs. The network bandwidth between machines is 25 Gbps. Table ~\ref{tab:throughput} shows the training throughput at epoch 20 when the number of GPUs increases to 12 times with the fixed mini-batch method and Dynamic SGD. The static baseline with 8 GPUs has a throughput of 4944 images per second.

\begin{table*}
\begin{center}
\begin{tabular}{l|l|l|l|l}
    \hline
     Method & Max Scale & ResNet-50 & MobileNet 1.0 & Inception V3 \\ \hline\hline
     Static Baseline & 1$\times$ & 76.91 & 72.16 & 77.78 \\ 
     Static Baseline & 12$\times$  & 76.34 & 72.02 & 77.02 \\ 
     Static Baseline & 16$\times$ & 76.08 & 71.82 & 76.30 \\ \hline
     Linear Scaling & 12$\times$ & 76.80 ($\pm$ 0.50) & 72.12 & 77.72 \\ 
     Dynamic SGD (ours) & 12$\times$ & 76.88 ($\pm$ 0.26) & 72.27 & 77.89 \\ \hline
     Linear Scaling & 16$\times$ & 76.10 ($\pm$ 0.33) & 71.56  & 76.88 \\ 
     Dynamic SGD (ours) & 16$\times$ & 76.65 ($\pm$ 0.56) & 71.99 & 77.61 \\ \hline
\end{tabular}
\end{center}
\caption{Top-1 accuracy for three models on ImageNet 2012 validation set with randomized GPU configurations.
         Static Baselines get worse with larger mini-batch size. When the maximum scale of mini-batch size is 12 times,
         both linear scaling and Dynamic SGD work well. When the maximum scale is pushed to 16 times, linear scaling
         gets much less accurate while Dynamic SGD keeps a better performance.}
\label{tab:rand_step_cls}
\end{table*}

When the number of GPUs increases in the middle of the training, these two methods exhibit different scaling property. For the fixed mini-batch size method, the training throughput drops to 335 images per second, about 7\% of the static baseline throughput. Due to a smaller mini-batch size per GPU, less time is spent on computing the loss and gradient, and more time is spent on communication and synchronization between GPUs. In contrast, the Dynamic SGD method increases the mini-batch size as the number of GPUs increases, which leads to a higher computation-communication ratio. Increasing the number of GPUs to 12 times boosts the training throughput to 8.11 times.


\paragraph{Benchmark on Random Schedules. }


We further study a more general scenario in which the number of GPUs may change frequently to a random scale up to a certain limit.  
We simulate the environment with two configurations, where the minimum number of GPUs is 8, and the maximum number of GPUs is 96 (12 times) and 128 (16 times) respectively. Given the range, we randomly change the number of GPUs within the range every 5 epochs. We have performed experiments for three image classification networks: ResNet-50, MobileNet 1.0 and Inception V3 with an identical training setup. 
To have more robust results, we repeatedly generate the random schedules on ResNet-50 for 5 times and report the mean and standard deviation of the top-1 accuracy metric. All results are included in Table~\ref{tab:rand_step_cls}.


In Table~\ref{tab:rand_step_cls}, we fix the numbers of GPUs to 8, 96 and 128 during the training as baselines. For all three networks, the baselines with 96 GPUs loose accuracies, and with 128 GPUs drop  accuracies significantly. It can also be seen that MobileNet 1.0 is more robust to the number of GPUs than the other two networks.

For each experiment for the random schedules, we randomly generate a schedule for the number of GPUs and train both our method and the linear scaling method with it, so that the results are comparable. When the maximum number of GPUs is 96, we can see that both our method and linear scale are better than the baseline with 96 GPUs and they have similar performance. However if we further set the maximum number of GPUs to 128, we can see that our method is still close to our 8-GPU baseline while the linear scale method has a much worse performance.

Based on the experimental results, our method has a consistently good performance and is comparable to the 8-GPU baseline accuracy. The linear scaling method can keep the performance if the maximum number of GPUs is not too large, e.g. twelve times of the baseline in our experiments, and its performance drops if we push the maximum number further.

\subsection{Elastic Training for Object Detection}

We go beyond the image classification and study elastic training on the object detection task. We first describe the implementation detail of the baseline network and the simulation procedure. We then study the performance with random schedules.

\paragraph{Implementation Detail} We train Single Shot Multi-box Object Detector network (SSD) \cite{liu2016ssd} on the MS-COCO dataset~\cite{lin2014microsoft}. We use \textit{train2017} imageset for training, and \textit{val2017} imageset for validation. We adopt ResNet-50 as the backbone and use $512 \times 512$ as the input image size. We use extensive data augmentation, hard negative sampling mining and normalize the loss by valid object count\cite{liu2016ssd}. The baseline network is trained with $0.008$ learning rate, mini-batch size 256 on 32 GPUs, with a cosine learning rate schedule and gradual learning rate warmup in the first 5 epochs\cite{zhang2019bag}.

\begin{table}[t]
\begin{center}
\begin{adjustbox}{max width=\linewidth}
\begin{tabular}{l|l|l}
\hline
Method & Max Scale & mAP \\ \hline\hline
Static Baseline      & 1$\times$  & 30.3\\
Static Baseline      & 12$\times$ & 30.0\\
Static Baseline      & 16$\times$ & 29.6\\ \hline
Linear Scaling       & 12$\times$ & 30.0\\
Dynamic SGD (ours)   & 12$\times$ & 30.1\\\hline
Linear Scaling       & 16$\times$ & 20.2\\
Dynamic SGD (ours)   & 16$\times$ & 29.6\\\hline

\end{tabular}
\end{adjustbox}
\end{center}
\caption{Mean average precision (mAP) for SSD network on the MS-COCO 2017 val set with randomized GPU configurations.  with randomized GPU configurations. We report mAP with IoU threshold 0.5:0.95, 0.5 and 0.75. When the maximum scale of mini-batch is 12 times, both linear scaling and Dynamic SGD perform relatively well. However, linear scaling method breaks down when the maximum scale is set to 16 times, while ours has similar performance compared to static baselines.}
\label{tab:ssd}
\end{table}

We randomly generate a schedule for the number of GPUs and train both our method and linear scaling with it. The number of GPUs changes every 5 epochs with a maximum scale of 12 times and 16 times respectively. When the maximum scale is 12 times, we see that both linear scaling method and ours performs relatively good compared to the static baseline. If the maximum scale is set to 16 times, we can see that our method is still close to our static baseline, while the linear scaling method fails to converge to a reasonable accuracy. 



\subsection{Elastic Training for Semantic Segmentation}

We also evaluate the Dynamic SGD method on semantic segmentation on ADE20K dataset~\cite{zhou2017scene}, which is a large scale scene parsing benchmark containing 20K/2K/3K images for training/validation and test set. We use Fully Convolutional Network (FCN)~\cite{long2015fully} as the baseline approach. Following the prior work~\cite{zhao2017pyramid}, we use the ResNet-50~\cite{he2016deep} as the base network and apply the dilation strategy at stage 3 \& 4, resulting in a stride-8 model. We use synchronized Batch Normalization\footnote{SyncBN is described in the appendix.}~\cite{Zhang_2018_CVPR} with a working batch size of 16. The baseline model is trained using a mini-batch size of 16, and base learning rate of 0.1 with cosine learning rate decay. For random schedule training, the training starts with 8 GPUs and the GPU scale is randomly chosen from 8 to maximum scales (12 times, 16 times) every 5 epochs. 

Pixel accuracy (pixAcc) and mean intersection of union (mIoU) are used as the evaluation metrics. The results are shown in Table~\ref{tab:fcn}. The static baseline 1$\times$ achieves 78.99\% pixAcc and 39.48\% mIoU. For random schedule, the proposed Dynamic SGD always outperform linear scaling method and achieves comparable performance with static baseline 1$\times$. 
For static baseline 12$\times$ and static baseline 16$\times$, FCN gets dramatically worse performance, which we believe is because the running statistics in batch normalization is not compensated for large mini-batch size. Different from SSD experiment where the batch normalization uses fixed batch statistics, FCN training employs Synchronized Batch Normalization. The moving averages of batch statistics are accumulated with momentum update, which is much slower than the update of network weights for large mini-batch size. This leads to degraded performance for static baseline. For the experiments using random schedule, the mini-batch size may not be always large and may alleviate the problem. Image classification training is not impacted by the batch normalization, because the training has larger number interactions per batch to accumulate batch statistics.

\begin{table}[t]
\begin{center}
\begin{adjustbox}{max width=\linewidth}
\begin{tabular}{l|l|l|l}
\hline
Method & Max Scale & pixAcc \% & mIoU \% \\ \hline\hline
Static Baseline      & 1$\times$  & 78.99 & 39.48  \\ 
Static Baseline      & 12$\times$ & 76.76 & 34.69  \\ 
Static Baseline      & 16$\times$ & 76.01 & 33.65  \\ \hline
Linear Scaling       & 12$\times$ & 78.94 & 39.16  \\
Dynamic SGD (ours) & 12$\times$ & 79.07 & 39.34  \\ \hline
Linear Scaling       & 16$\times$ & 79.01 & 38.90  \\
Dynamic SGD (ours) & 16$\times$ & 79.02 & 39.34  \\ \hline

\end{tabular}
\end{adjustbox}
\end{center}
\caption{Elastic Training results using random schedules on semantic segmentation,  showing pixAcc and mIoU of FCN on ADE20K validation set. 
The Dynamic SGD is compared with static baseline and linear scaling using $12\times$ and $16\times$ maximum GPU scales. }
\label{tab:fcn}
\end{table}






\section{Conclusion and Future Work} 
\label{sec:conclusion}


In this paper, we study the optimization difficulty in elastic distributed training, which is the fundamental bottleneck for dynamic scheduling in deep learning system. 
We find this difficulty is mainly because the momentum is not compensated for the changing mini-batch size. For this, we introduce Dynamic SGD to gradually adapt the momentum. 
The proposed Dynamic SGD has been evaluated on three major computer vision tasks: image classification, object detection and semantic segmentation. 
The experimental results demonstrate the proposed method stabilizes the training and achieves comparable model accuracy after convergence with single node training using the same training settings. 
The proposed Dynamic SGD method can also employ other momentum compensation strategies, which is discussed in the appendix. 



{\small
\bibliographystyle{ieee}
\bibliography{main}

\begin{thebibliography}{10}\itemsep=-1pt

\bibitem{allen2017katyusha}
Z.~Allen-Zhu.
\newblock Katyusha: The first direct acceleration of stochastic gradient
  methods.
\newblock {\em The Journal of Machine Learning Research}, 18(1):8194--8244,
  2017.

\bibitem{chen2016revisiting}
J.~Chen, X.~Pan, R.~Monga, S.~Bengio, and R.~Jozefowicz.
\newblock Revisiting distributed synchronous sgd.
\newblock {\em arXiv preprint arXiv:1604.00981}, 2016.

\bibitem{chen2018deeplab}
L.-C. Chen, G.~Papandreou, I.~Kokkinos, K.~Murphy, and A.~L. Yuille.
\newblock Deeplab: Semantic image segmentation with deep convolutional nets,
  atrous convolution, and fully connected crfs.
\newblock {\em IEEE transactions on pattern analysis and machine intelligence},
  40(4):834--848, 2018.

\bibitem{chen2015mxnet}
T.~Chen, M.~Li, Y.~Li, M.~Lin, N.~Wang, M.~Wang, T.~Xiao, B.~Xu, C.~Zhang, and
  Z.~Zhang.
\newblock Mxnet: A flexible and efficient machine learning library for
  heterogeneous distributed systems.
\newblock {\em arXiv preprint arXiv:1512.01274}, 2015.

\bibitem{NIPS2012_4687}
J.~Dean, G.~Corrado, R.~Monga, K.~Chen, M.~Devin, M.~Mao, A.~Senior, P.~Tucker,
  K.~Yang, Q.~V. Le, et~al.
\newblock Large scale distributed deep networks.
\newblock In {\em Advances in neural information processing systems}, pages
  1223--1231, 2012.

\bibitem{deng2009imagenet}
J.~Deng, W.~Dong, R.~Socher, L.-J. Li, K.~Li, and L.~Fei-Fei.
\newblock Imagenet: A large-scale hierarchical image database.
\newblock In {\em Computer Vision and Pattern Recognition, 2009. CVPR 2009.
  IEEE Conference on}, pages 248--255. Ieee, 2009.

\bibitem{devarakonda2017adabatch}
A.~Devarakonda, M.~Naumov, and M.~Garland.
\newblock Adabatch: Adaptive batch sizes for training deep neural networks.
\newblock {\em arXiv preprint arXiv:1712.02029}, 2017.

\bibitem{feichtenhofer2018slowfast}
C.~Feichtenhofer, H.~Fan, J.~Malik, and K.~He.
\newblock Slowfast networks for video recognition.
\newblock {\em arXiv preprint arXiv:1812.03982}, 2018.

\bibitem{ghadimi2013stochastic}
S.~Ghadimi and G.~Lan.
\newblock Stochastic first-and zeroth-order methods for nonconvex stochastic
  programming.
\newblock {\em SIAM Journal on Optimization}, 23(4):2341--2368, 2013.

\bibitem{google}
Google.
\newblock Preemptible vm, Jan 2018.

\bibitem{goyal2017accurate}
P.~Goyal, P.~Doll{\'a}r, R.~Girshick, P.~Noordhuis, L.~Wesolowski, A.~Kyrola,
  A.~Tulloch, Y.~Jia, and K.~He.
\newblock Accurate, large minibatch sgd: training imagenet in 1 hour.
\newblock {\em arXiv preprint arXiv:1706.02677}, 2017.

\bibitem{DBLP:journals/corr/GoyalDGNWKTJH17}
P.~Goyal, P.~Doll{\'{a}}r, R.~B. Girshick, P.~Noordhuis, L.~Wesolowski,
  A.~Kyrola, A.~Tulloch, Y.~Jia, and K.~He.
\newblock Accurate, large minibatch {SGD:} training imagenet in 1 hour.
\newblock {\em CoRR}, abs/1706.02677, 2017.

\bibitem{he2016deep}
K.~He, X.~Zhang, S.~Ren, and J.~Sun.
\newblock Deep residual learning for image recognition.
\newblock In {\em Proceedings of the IEEE conference on computer vision and
  pattern recognition}, pages 770--778, 2016.

\bibitem{hoffer2017train}
E.~Hoffer, I.~Hubara, and D.~Soudry.
\newblock Train longer, generalize better: closing the generalization gap in
  large batch training of neural networks.
\newblock In {\em Advances in Neural Information Processing Systems}, pages
  1731--1741, 2017.

\bibitem{howard2017mobilenets}
A.~G. Howard, M.~Zhu, B.~Chen, D.~Kalenichenko, W.~Wang, T.~Weyand,
  M.~Andreetto, and H.~Adam.
\newblock Mobilenets: Efficient convolutional neural networks for mobile vision
  applications.
\newblock {\em arXiv preprint arXiv:1704.04861}, 2017.

\bibitem{ioffe2015batch}
S.~Ioffe and C.~Szegedy.
\newblock Batch normalization: Accelerating deep network training by reducing
  internal covariate shift.
\newblock {\em arXiv preprint arXiv:1502.03167}, 2015.

\bibitem{jastrzkebski2017three}
S.~Jastrz{\k{e}}bski, Z.~Kenton, D.~Arpit, N.~Ballas, A.~Fischer, Y.~Bengio,
  and A.~Storkey.
\newblock Three factors influencing minima in sgd.
\newblock {\em arXiv preprint arXiv:1711.04623}, 2017.

\bibitem{jia2018highly}
X.~Jia, S.~Song, W.~He, Y.~Wang, H.~Rong, F.~Zhou, L.~Xie, Z.~Guo, Y.~Yang,
  L.~Yu, et~al.
\newblock Highly scalable deep learning training system with mixed-precision:
  Training imagenet in four minutes.
\newblock {\em arXiv preprint arXiv:1807.11205}, 2018.

\bibitem{jin2017accelerated}
C.~Jin, P.~Netrapalli, and M.~I. Jordan.
\newblock Accelerated gradient descent escapes saddle points faster than
  gradient descent.
\newblock {\em arXiv preprint arXiv:1711.10456}, 2017.

\bibitem{kingma2014adam}
D.~P. Kingma and J.~Ba.
\newblock Adam: A method for stochastic optimization.
\newblock {\em arXiv preprint arXiv:1412.6980}, 2014.

\bibitem{konevcny2016mini}
J.~Kone{\v{c}}n{\`y}, J.~Liu, P.~Richt{\'a}rik, and M.~Tak{\'a}{\v{c}}.
\newblock Mini-batch semi-stochastic gradient descent in the proximal setting.
\newblock {\em IEEE Journal of Selected Topics in Signal Processing},
  10(2):242--255, 2016.

\bibitem{lei2018tvqa}
J.~Lei, L.~Yu, M.~Bansal, and T.~L. Berg.
\newblock Tvqa: Localized, compositional video question answering.
\newblock {\em arXiv preprint arXiv:1809.01696}, 2018.

\bibitem{li2017scaling}
M.~Li.
\newblock {\em Scaling Distributed Machine Learning with System and Algorithm
  Co-design}.
\newblock PhD thesis, PhD thesis, Carnegie Mellon University, 2017.

\bibitem{li2018low}
Y.~Li, J.~Shi, and D.~Lin.
\newblock Low-latency video semantic segmentation.
\newblock In {\em Proceedings of the IEEE Conference on Computer Vision and
  Pattern Recognition}, pages 5997--6005, 2018.

\bibitem{lin2014microsoft}
T.-Y. Lin, M.~Maire, S.~Belongie, J.~Hays, P.~Perona, D.~Ramanan,
  P.~Doll{\'a}r, and C.~L. Zitnick.
\newblock Microsoft coco: Common objects in context.
\newblock In {\em European conference on computer vision}, pages 740--755.
  Springer, 2014.

\bibitem{liu2016ssd}
W.~Liu, D.~Anguelov, D.~Erhan, C.~Szegedy, S.~Reed, C.-Y. Fu, and A.~C. Berg.
\newblock Ssd: Single shot multibox detector.
\newblock In {\em European conference on computer vision}, pages 21--37.
  Springer, 2016.

\bibitem{long2015fully}
J.~Long, E.~Shelhamer, and T.~Darrell.
\newblock Fully convolutional networks for semantic segmentation.
\newblock In {\em Proceedings of the IEEE conference on computer vision and
  pattern recognition}, pages 3431--3440, 2015.

\bibitem{DBLP:journals/corr/LoshchilovH16a}
I.~Loshchilov and F.~Hutter.
\newblock {SGDR:} stochastic gradient descent with restarts.
\newblock {\em CoRR}, abs/1608.03983, 2016.

\bibitem{DBLP:journals/corr/abs-1812-06162}
S.~McCandlish, J.~Kaplan, D.~Amodei, and O.~D. Team.
\newblock An empirical model of large-batch training.
\newblock {\em CoRR}, abs/1812.06162, 2018.

\bibitem{mccandlish2018empirical}
S.~McCandlish, J.~Kaplan, D.~Amodei, and O.~D. Team.
\newblock An empirical model of large-batch training.
\newblock {\em arXiv preprint arXiv:1812.06162}, 2018.

\bibitem{azure}
Microsoft.
\newblock Azure low priority vm, April 2018.

\bibitem{nemirovski2009robust}
A.~Nemirovski, A.~Juditsky, G.~Lan, and A.~Shapiro.
\newblock Robust stochastic approximation approach to stochastic programming.
\newblock {\em SIAM Journal on optimization}, 19(4):1574--1609, 2009.

\bibitem{nesterov1983method}
Y.~E. Nesterov.
\newblock A method for solving the convex programming problem with convergence
  rate o (1/k\^{} 2).
\newblock In {\em Dokl. akad. nauk Sssr}, volume 269, pages 543--547, 1983.

\bibitem{polyak1964some}
B.~T. Polyak.
\newblock Some methods of speeding up the convergence of iteration methods.
\newblock {\em USSR Computational Mathematics and Mathematical Physics},
  4(5):1--17, 1964.

\bibitem{qian1999momentum}
N.~Qian.
\newblock On the momentum term in gradient descent learning algorithms.
\newblock {\em Neural networks}, 12(1):145--151, 1999.

\bibitem{reddi2015variance}
S.~J. Reddi, A.~Hefny, S.~Sra, B.~Poczos, and A.~J. Smola.
\newblock On variance reduction in stochastic gradient descent and its
  asynchronous variants.
\newblock In {\em Advances in Neural Information Processing Systems}, pages
  2647--2655, 2015.

\bibitem{ren2015faster}
S.~Ren, K.~He, R.~Girshick, and J.~Sun.
\newblock Faster r-cnn: Towards real-time object detection with region proposal
  networks.
\newblock In {\em Advances in neural information processing systems}, pages
  91--99, 2015.

\bibitem{robbins1951stochastic}
H.~Robbins and S.~Monro.
\newblock A stochastic approximation method.
\newblock {\em The annals of mathematical statistics}, pages 400--407, 1951.

\bibitem{ec2spot}
A.~W. Services.
\newblock Amazon ec2 spot instances, August 2018.

\bibitem{smith2017don}
S.~L. Smith, P.-J. Kindermans, and Q.~V. Le.
\newblock Don't decay the learning rate, increase the batch size.
\newblock {\em arXiv preprint arXiv:1711.00489}, 2017.

\bibitem{szegedy2015going}
C.~Szegedy, W.~Liu, Y.~Jia, P.~Sermanet, S.~Reed, D.~Anguelov, D.~Erhan,
  V.~Vanhoucke, and A.~Rabinovich.
\newblock Going deeper with convolutions.
\newblock In {\em Proceedings of the IEEE conference on computer vision and
  pattern recognition}, pages 1--9, 2015.

\bibitem{DBLP:journals/corr/SzegedyVISW15}
C.~Szegedy, V.~Vanhoucke, S.~Ioffe, J.~Shlens, and Z.~Wojna.
\newblock Rethinking the inception architecture for computer vision.
\newblock {\em CoRR}, abs/1512.00567, 2015.

\bibitem{tapaswi2016movieqa}
M.~Tapaswi, Y.~Zhu, R.~Stiefelhagen, A.~Torralba, R.~Urtasun, and S.~Fidler.
\newblock Movieqa: Understanding stories in movies through question-answering.
\newblock In {\em Proceedings of the IEEE conference on computer vision and
  pattern recognition}, pages 4631--4640, 2016.

\bibitem{voigtlaender2019feelvos}
P.~Voigtlaender, Y.~Chai, F.~Schroff, H.~Adam, B.~Leibe, and L.-C. Chen.
\newblock Feelvos: Fast end-to-end embedding learning for video object
  segmentation.
\newblock {\em arXiv preprint arXiv:1902.09513}, 2019.

\bibitem{wang2018non}
X.~Wang, R.~Girshick, A.~Gupta, and K.~He.
\newblock Non-local neural networks.
\newblock In {\em Proceedings of the IEEE Conference on Computer Vision and
  Pattern Recognition}, pages 7794--7803, 2018.

\bibitem{xie2018bag}
J.~Xie, T.~He, Z.~Zhang, H.~Zhang, Z.~Zhang, and M.~Li.
\newblock Bag of tricks for image classification with convolutional neural
  networks.
\newblock {\em arXiv preprint arXiv:1812.01187}, 2018.

\bibitem{you2017large}
Y.~You, I.~Gitman, and B.~Ginsburg.
\newblock Large batch training of convolutional networks. arxiv preprint.
\newblock {\em arXiv preprint arXiv:1708.03888}, 2017.

\bibitem{Zhang_2018_CVPR}
H.~Zhang, K.~Dana, J.~Shi, Z.~Zhang, X.~Wang, A.~Tyagi, and A.~Agrawal.
\newblock Context encoding for semantic segmentation.
\newblock In {\em The IEEE Conference on Computer Vision and Pattern
  Recognition (CVPR)}, June 2018.

\bibitem{zhang2015staleness}
W.~Zhang, S.~Gupta, X.~Lian, and J.~Liu.
\newblock Staleness-aware async-sgd for distributed deep learning.
\newblock {\em arXiv preprint arXiv:1511.05950}, 2015.

\bibitem{zhang2019bag}
Z.~Zhang, T.~He, H.~Zhang, Z.~Zhang, J.~Xie, and M.~Li.
\newblock Bag of freebies for training object detection neural networks.
\newblock {\em arXiv preprint arXiv:1902.04103}, 2019.

\bibitem{zhao2017pyramid}
H.~Zhao, J.~Shi, X.~Qi, X.~Wang, and J.~Jia.
\newblock Pyramid scene parsing network.
\newblock In {\em Computer Vision and Pattern Recognition (CVPR), 2017 IEEE
  Conference on}, pages 6230--6239. IEEE, 2017.

\bibitem{zheng2017asynchronous}
S.~Zheng, Q.~Meng, T.~Wang, W.~Chen, N.~Yu, Z.-M. Ma, and T.-Y. Liu.
\newblock Asynchronous stochastic gradient descent with delay compensation.
\newblock In {\em Proceedings of the 34th International Conference on Machine
  Learning-Volume 70}, pages 4120--4129. JMLR. org, 2017.

\bibitem{zhou2017scene}
B.~Zhou, H.~Zhao, X.~Puig, S.~Fidler, A.~Barriuso, and A.~Torralba.
\newblock Scene parsing through ade20k dataset.
\newblock In {\em Proceedings of the IEEE Conference on Computer Vision and
  Pattern Recognition}, 2017.

\end{thebibliography}
}
\clearpage

\appendix
\section*{Appendix}


\section{Linear Scaling with Increased \#GPUs}

Weight decay is a common practice to regularize neural networks. The experimental results at Section 5.1 do not apply weight decay biases as well as $\gamma$ and $\beta$ in batch normalization layers. In this section, we study the case where weight decay is applied. Here we also adopt ResNet-50~\cite{he2016deep} and train it using the same implementation as described in Section 5.1. The only difference is that here weight decay is applied to biases as well as $\gamma$ and $\beta$ in batch normalization layers. We also simulate the gradient of a large mini-batch by accumulating gradients from multiple small mini-batches before applying the parameter update.

\begin{figure}
\centering
\subfloat[Increase the mini-batch size to 12 times at epoch 20.]{
    \includegraphics[width=0.9\linewidth]{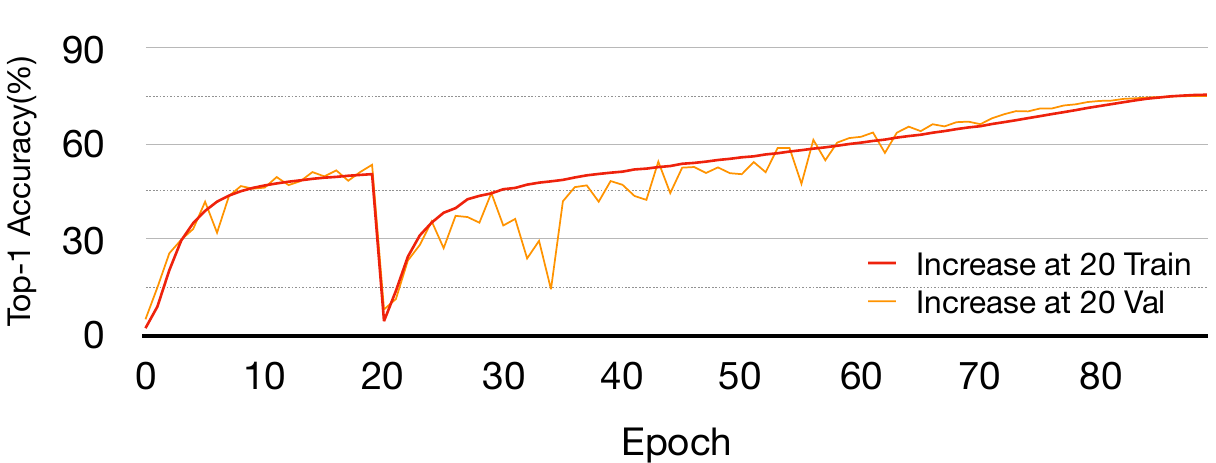}
}\\
\subfloat[Decrease the mini-batch size to $\frac{1}{12}$ at epoch 20.]{
    \includegraphics[width=0.9\linewidth]{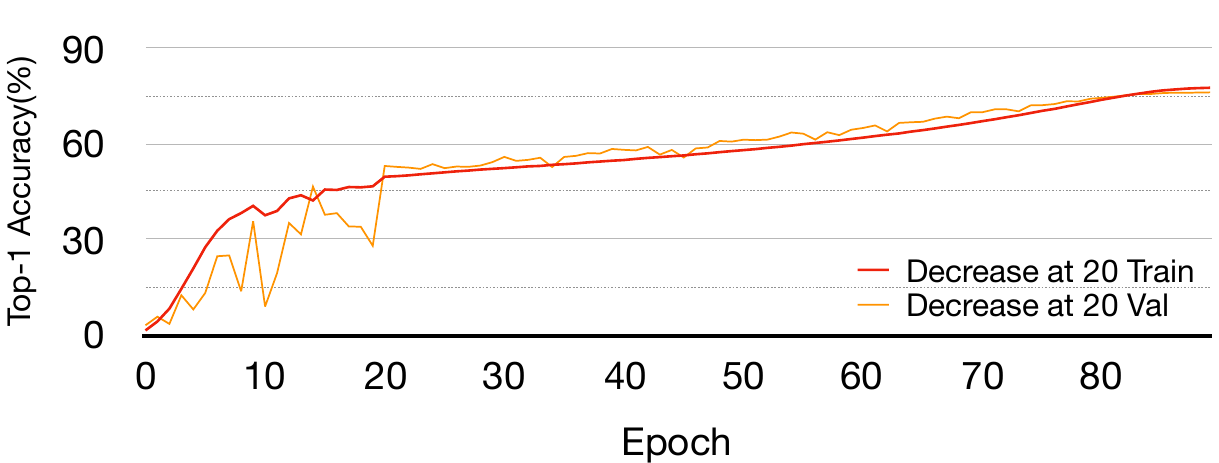}
}
\caption{Top-1 accuracy on ImageNet validation set using ResNet-50, when weight decay is applied. 
Influence of sudden change in mini-batch size using linear scaling approach. 
We use the 1K as the base mini-batch size. We find that the network training is relatively sensitive to increasing mini-batch size, but robust to decreasing.}

\label{fig:spike_dump_has_wd}
\end{figure}

We study the influence of a sudden change of number of GPUs at epoch 20. We train the network with two configurations: 1) training starts from 8 GPUs and then increase the number of GPUs to 96 at epoch 20, and 2) training starts from 96 GPUs and then decrease the number of GPUs to 8 at epoch 20. In both configurations we scale up or down the learning rate linearly with the number of GPUs. The results are included in Figure~\ref{fig:spike_dump_has_wd}. 

Figure~\ref{fig:spike_dump_has_wd} shows that by increasing the number of GPUs to 96, the training curve has a sharp drop, indicating the training process is drastically disrupted at epoch 20. On the other hand, decreasing the number of GPUs has no visible effect. This leads to a more through study for the case where the number of GPUs is suddenly increased.

\begin{table}
\begin{center}
\begin{tabular}{l|l|l|l}
    \hline
     Method   & Scale & Epoch 20 & Epoch 70  \\ \hline\hline
     Static Baseline & 1$\times$  & \multicolumn{2}{c}{76.89} \\ 
     Static Baseline & 12$\times$ & \multicolumn{2}{c}{76.27} \\ \hline
     Linear Scaling   & 12$\times$ & 75.03 & 76.43 \\ 
     Dynamic SGD (ours)   & 12$\times$ & 76.47 & 76.78 \\ \hline

\end{tabular}
\end{center}
\caption{Performance of different methods when increasing the number of GPUs at early or late stage of the training process, when weight decay is present.
Our method is usually not affected by the sudden increment and outperforms the linear method. Linear scaling breaks down if the change happens at epoch 20, 1.86\% worse than the static 1 $\times$ baseline.}
\label{tab:spike_12_wd}
\end{table}

Table~\ref{tab:spike_12_wd} shows that our Dynamic SGD method has a consistent performance against the sudden change at both early and late stage. 
The linear scaling method of learning rate works well if we increase the number of GPUs at a later epoch. If the increase happens at epoch 20, the linear scaling method perform much worse than the static baseline 12 $\times$. Meanwhile our method performs consistently well on increasing number of GPUs at both early and later epochs.

\section{Discussions}

This section provides further theoretical discussions on (1) our choice of linear relations between the learning rate and the number of machines and (2) our decision to warm-up linear rate when the number of machines changes. The discussions extend the main text in terms of technical details and generality.

\subsection{Linear LR Scaling in Convex Optimization}
\label{app:1}

We briefly mention the reasons with a simplified convex problem, similar to \cite{mccandlish2018empirical}.
First, optimal learning rate in gradient descent is chosen by optimizing a conservative upper bound,
\begin{align*}
    L(w_t-\eta_{t+1}\nabla l(w_t))&\leq L(w_t) -\eta_{t+1} \nabla l(w_t)^\top\nabla L(w_t) 
    \\
    &+ \frac{C\eta_{t+1}^2}{2}\nabla l(w_t)^\top \nabla l(w_t),
\end{align*}
where $C$ is the Lipschitz constant which bounds all smaller terms in the  Taylor expansion.
If we additionally assume an i.i.d. Gaussian noise of the stochastic gradient,
\begin{equation*}
    \nabla l(w_t)=\nabla L(w_t)+\varepsilon_t,
\end{equation*}
where $\varepsilon_t\sim\mathcal{N}(0,\sigma_k^2)$ is the noise of the average gradient estimation. Here $\sigma_k^2$ is the variance of $\varepsilon$ from $k$ machines, which decreases as  $k$ increases, i.e., $\sigma_k^2=\frac{\sigma_1^2}{k}$.
The expected loss becomes,
\begin{align*}
    \mathbb{E} L(w_t-\eta_{t+1}\nabla l(w_t))
    &\leq L(w_t) - \eta_{t+1} \nabla L(w_t)^\top\nabla L(w_t)
    \nonumber\\
    &+ \frac{C\eta_{t+1}^2}{2}(\nabla L(w_t)^\top\nabla L(w_t) + \sigma_k^2).
\end{align*}
Define $G^2=\nabla L(w_t)^\top\nabla L(w_t)$, the optimal choice is
$
    \eta_{t+1}=\frac{1}{C}\frac{G^2}{G^2+\sigma_k^2}=\frac{kG^2}{C(kG^2+\sigma_1^2)}
    \approx \frac{kG^2}{C\sigma_1^2},
$
where the last approximation assumes  $kG^2\ll \sigma_1^2$, common in SGD with many mini-batches. In this way, we show that $\eta_{t+1}$ grows linearly with machine size $k$.



\subsection{Linear LR Scaling in Non-convex Optimization}
\label{sec:theory}

This section extends the previous discussions to non-convex stochastic optimization.
Our theoretical analysis also extends \cite{ghadimi2013stochastic}, which showed that non-convex stochastic gradient descent converges to a stationary point at a near-optimal rate of $O(\nicefrac{1}{\sqrt{T}})$.

\renewcommand{\theta}{w}

\begin{assumption}
\label{assumption}
Let $L_{\Delta}=L(\theta_0)-\min L(\theta)<\infty$ be the total range of the objective function,
which is, without loss of generality, finite.
Suppose $L(\theta)$ has bounded second-order gradient, $-CI\preceq\nabla^2 f(\theta)\preceq CI$.
For iteration $t$, let the number of machines be $k_t$ such that $1\leq k_t \leq K$, where $K$ is the maximum machine size, and the
variance of the stochastic gradient $\nabla l_t(\theta)$ be
$\mathbb{E}[\|\nabla l_t(\theta) - \nabla L_t(\theta)\|^2]\leq \frac{\sigma_1^2}{k_t}$,
which is the same assumption we used in the appendix of the main text.
\end{assumption}

We aim to analyze the convergence of a dynamic learning rate scaling rule, where at iteration $t$, the step size is $\eta_t=k_t^\beta\eta_0$ for some constant $\beta\geq0$.
Choosing $\beta=0$ yields a constant learning rate, which may be suboptimal.
Define constants $T_0=\frac{2CKL_{\Delta}}{\sigma_1^2}$, $C_1=\sqrt{\frac{2L_{\Delta}}{C\sigma_1^2}}$ and $C_2=\sqrt{2C\sigma_1^2L_{\Delta}}$, which do not depend on the choice of the step sizes $\eta_t$.

\begin{thm}[Convergence with Dynamic Machine Size]
\label{thm:convergence}
Under Assumption~\ref{assumption} with dynamic machine sizes $k_t\forall t$,
if we adopt a strategy that sets the step size to $\eta_t=k_t^\beta\eta_0$, where $\beta\geq0$ is a predefined constant, and have sufficiently many gradient steps,
$T ( > T_0)$,
then choosing
\begin{equation}
 \label{eq:condition}
    \eta_t=
    \frac{C_1 k_t^\beta}{\sqrt{\sum_{t=1}^T k_t^{2\beta-1}}}
\end{equation}
guarantees at least one solution at the stationary point whose expected gradient is at most:
\begin{equation}
\label{eq:convergence}
    \min_{t\leq T}\mathbb{E}[\|\nabla L(\theta_t)\|^2]
    \leq
    C_2
    \frac{
    \sqrt{\sum_{t=1}^T k_t^{2\beta-1}}
    }
    {\sum_{t=1}^T k_t^\beta}.
\end{equation}
\end{thm}



\begin{remark}
Equation \eqref{eq:convergence} converges to zero 
for any learning rate scaling rule $\eta_t=k_t^\beta\eta_0$, as long as $\beta>0$ and Equation~\eqref{eq:condition} is satisfied.
For example, with a fixed step size $\eta_t=\eta_0 (<\frac{1}{C})$, Theorem~\ref{thm:convergence} reduces to \cite{nemirovski2009robust}.
\end{remark}

\begin{remark}
Equation \eqref{eq:convergence} suggests that the upper bound of the convergence rate becomes optimal when the learning rate is $\eta_t=k_t\eta_0$ with $\beta=1$.
To see how, notice that the right-hand side of \eqref{eq:convergence} can be relaxed using Cauchy's inequality as
\begin{equation}
    \frac{\sqrt{\sum_{t=1}^T k_t^{2\beta-1}}}
    {\sum_{t=1}^T k_t^\beta}
    \geq
    \frac{1}
    {\sqrt{\sum_{t=1}^T k_t}},
\end{equation}
which is tight when $\beta=1$.
The convergence rate is comparable with $O(\nicefrac{1}{\sqrt{T}})$ when the machine size is fixed.
\end{remark}

\begin{remark}
In practice, the number of steps $T$ and the bound on the second-order gradients $C$ may not be known ahead of time.
To adapt to any $C$ and $T$, the learning rate is often also decreasing over time, i.e. $\eta_t=f(t)k_t^\beta\eta_0$.
In our experiments, the learning rate follows a cosine function $f(t)=\cos(\frac{\pi}{2}\frac{t}{90})$, which is monotonically decreasing over time $(t<90)$.
Theoretical analysis \cite{nemirovski2009robust} on a similar decreasing rate $f(t)=\nicefrac{1}{t}$
shows that it has a suboptimal convergence rate of $O(\nicefrac{1}{\log(T)})$, but is more generalizable to unknown conditions.
\end{remark}


\begin{proof}[Proof of Theorem~\ref{thm:convergence}]
The original proof in \cite{ghadimi2013stochastic} focuses on the quantity
\begin{equation}
    \mathcal{E} = \frac{1}{\sum_{t=1}^T\eta_t}\sum_{t=1}^T\eta_t\mathbb{E}[\|\nabla L(\theta_t)\|^2],
\end{equation}
which is an upper bound of the left-hand side of \eqref{eq:convergence}, because the weighted average of a sample is greater than its minimum value.
To bound $\mathcal{E}$, \cite{ghadimi2013stochastic} then expands the noise condition in Assumption~\ref{assumption} to obtain
\begin{equation}
 \label{eq:initial-inequality}
    \mathcal{E}
    \leq
    \frac{2L_{\Delta}}{\sum_t^T\eta_t}
    +
    \frac{\sum_t^T C \eta_t^2 \sigma_t^2}{\sum_t^T \eta_t},
\end{equation}
where the first term connects the changes in the function values with noiseless gradients and the second term reflects the noise introduced by using stochastic gradients.
In our case, the stochastic gradient noise at iteration $t$ is $\sigma_t^2\leq \frac{\sigma_1^2}{k_t}$.
We further plug in the learning rate $\eta_t=k_t^\beta\eta_0$ to rewrite \eqref{eq:initial-inequality} as
\begin{equation}
  \mathcal{E}
  \leq
    \frac{2L_{\Delta}}{\eta_0\sum_t^T k_t^\beta}
    +
    \frac{\eta_0 C\sigma_1^2\sum_t^T k_t^{2\beta-1}}{\sum_t^T k_t^\beta}.
    \label{eq:choose-eta-0}
\end{equation}
Since \eqref{eq:choose-eta-0} works for any choice of $\eta_0$, Cauchy's inequality suggests the optimal $\eta_0$ such that the two terms on the right-hand side equal, i.e., $\eta_0^2=\frac{2L_\Delta}{C\sigma_1^2}\frac{1}{\sum_{t=1}^T k_t^{2\beta-1}}=\frac{C_1^2}{\sum_{t=1}^T k_t^{2\beta-1}}$. Straightforward calculation recovers \eqref{eq:condition} and \eqref{eq:convergence}.
\label{eq:proof}
\end{proof}

\begin{figure}
\centering
\includegraphics[width=0.8\linewidth]{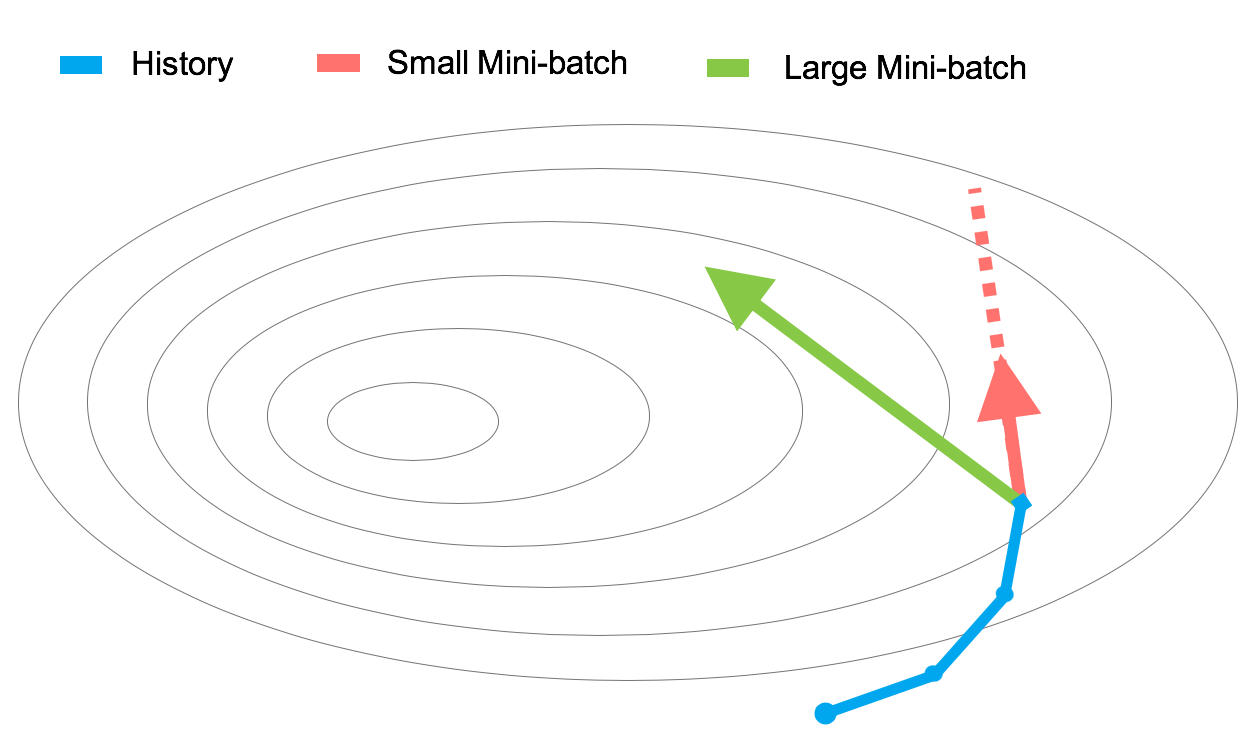}
\caption{Gradient of larger mini-batch usually has better approximation of the true gradient. We assume the momentum state has a similar direction with gradient as shown using red dashed line. In momentum SGD, increasing the mini-batch size during the training without compensating for the history momentum.}
\label{fig:spike_compare}
\end{figure}

\subsection{Momentum and Learning Rate Warm-up}

Momentum is helpful in convex problems with gradient descent by improving the condition number of the objective function \cite{polyak1964some,nesterov1983method}.
Momentum also helps in non-convex problems with gradient descent by escaping the saddle points \cite{jin2017accelerated}.
While momentum in SGD can be prone to error accumulation \cite{konevcny2016mini}, their adoption usually helps in practice, especially when the gradient noise is small in large batch training \cite{allen2017katyusha}.

The challenge we face in a dynamic environment is that the variance of the momentum terms does not immediately change after the machine size changes.
This violates Assumption~\ref{assumption} in Theorem~\ref{thm:convergence} that the variance in the descent direction is inversely proportional to the machine size, e.g., in the case of SGD without momentum.
However, the variance of the momentum will monotonically decrease until $O(\nicefrac{1}{1-\mu})$ number of epochs, where $\mu$ is the momentum decay rate, e.g., $\mu=0.9$.
We therefore apply a learning rate warm-up strategy such that the variance in the descent direction, which is the product of the variance of the momentum itself and the step size, maintains a relatively stable value.
After the learning rate warm-up, the variance of the descent direction will also be inversely proportional to the machine size and the same linear scale would apply.

\subsection{Thoughts on Momentum Compensation}

We adopt a simple yet effective strategy in this paper to gradually increase the learning rate for smooth adaption of momentum state, which works well empirically. 
However, the proposed Dynamic SGD can also employ other momentum compensation strategies. 
Inspired by the physical analogy of Newtonian particles in a conservative force field in Qian~\cite{qian1999momentum}, we can consider the loss $L(w)$ as the energy function, $\mu$ as the system friction coefficient,  and the $v_t$ as the velocity in Equation 4 and the mini-batch gradient $\sum_{i=1}^B\nabla l(w_t, x_i)$ as the acceleration. 
Therefore, the ``velocity'' can be dynamically adjusted by the system without introducing extra hyper-parameters, such as momentum compensation factor $\gamma_t$. The parameter update is given by:
\begin{equation}
\begin{split}
    &v_{t+1} = \mu v_t + \hat{\eta} \sum_{i=1}^B\nabla l(w_t, x_i) \\
    &w_{t+1} = w_t - v_{t+1} ,
\end{split}
\end{equation}

where $\hat{\eta}$ is the step size, which is different from learning rate as it is not coupled with mini-batch size. 
Comparing to the method proposed in Section 3.2, 1) this strategy does not introduce extra hyper-parameter, 2) does not need warm-up-like gradually adaption, 3) is generalized both increasing and decreasing mini-batch situations.
The experimental analysis for different momentum compensation will be addressed in our future work. 

\section{Batch Normalization with Data Parallel}
\label{app:BN}
The batch normalization layer~\cite{ioffe2015batch} normalizes the data within a mini-batch, which makes the network less sensitive to the initialization and allows larger learning rate training. Batch normalization is commonly used in modern deep convolution neural networks. 
Standard implementation of batch normalization in data parallel training normalizes the data within each worker or GPU. 

Denotes the per-worker batch size as $S$, and the inputs to a batch normalization layer as $x_1, \ldots, x_S$. 
The mean $\mu$ and variance $\sigma^2$ for the inputs are given by $\mu = \frac1S \sum_{i = 1}^S x_i$ and $\sigma^2 = \frac1S \sum_{i=1}^S (x_i - \mu)^2$. 
The outputs $y_i \in \{y_1,\ldots, y_S\}$ is given by:

\begin{equation}
y_i = \gamma \frac{x_i - \mu}{\sqrt{\sigma^2 + \epsilon}} + \beta ,
\label{eq:bn}
\end{equation}

where  $\gamma$ and $\beta$ are the scale and shift parameters, and $\epsilon$ is a small constant to avoid extreme values.
As can be seen, when changing the per-worker batch size $S$, the batch normalization layer standardizes the inputs differently. 

Different than the standard implementation of BN, Synchronized Batch Normalization (SyncBN)~\cite{Zhang_2018_CVPR} normalizes the data using mean and variance calculated across multiple worker/GPUs, and synchronized their gradients during the backward pass. 
SyncBN is usually used in semantic segmentation task, where the per-worker batch size is often small. It is not suitable for image classification, because the synchronization becomes a big communication overhead to the training. 

\end{document}